\def\year{2020}\relax
\documentclass[letterpaper]{article} 
\usepackage{aaai20}  
\usepackage{times}  
\usepackage{helvet} 
\usepackage{courier}  
\usepackage[hyphens]{url}  
\usepackage{graphicx} 
\usepackage{graphicx}  
\usepackage{amsmath}
\usepackage{amssymb}
\usepackage{multirow}
\usepackage{subfigure}
\usepackage{chngpage}
\usepackage{xcolor,colortbl}
\usepackage{bbold}
\usepackage{dsfont}
\usepackage{booktabs}
\newcommand{\myparagraph}[1]{{\vspace{0.5em} \noindent \bf #1}}
\newcommand{\etal}{\textit{et al}.}

\newcommand{\eg}{\textit{e}.\textit{g}.}

\title{Task-Aware Monocular Depth Estimation for 3D Object Detection}

\author{Xinlong Wang$^{1}$\thanks{This work is done when Xinlong Wang is an intern
at Bytedance AI Lab.},
Wei Yin$^{1}$,
Tao Kong$^{2}$,
Yuning Jiang$^{2}$,
Lei Li$^{2}$, 
Chunhua Shen$^{1}$ \\
\\
$^1$The University of Adelaide, Australia  ~ ~ ~ 
$^2$Bytedance AI Lab  \\
\tt\small \{xinlong.wang, wei.yin, chunhua.shen\}@adelaide.edu.au,\\ \tt\small taokongcn@gmail.com, \{jiangyuning, lileilab\}@bytedance.com
}

\begin{document}

\maketitle

\begin{abstract}
   Monocular depth estimation enables 3D perception from a single 2D image, thus attracting much research attention for years. 
Almost all 
   methods treat foreground and background regions 
   (``things and stuff'')
   in an image equally.
   However,  not all pixels are equal. 
   Depth of foreground objects plays a crucial role in 3D object recognition and localization.
   To date how to boost the depth prediction accuracy of foreground objects is rarely discussed.
   In this paper, we first analyse the data distributions and interaction of foreground and background, then propose the foreground-background separated monocular depth estimation (ForeSeE) method, to estimate the foreground and background depth using separate optimization objectives and  decoders.
   Our method 
   significantly 
   improves the depth estimation performance on foreground objects.
   Applying ForeSeE to 3D object detection, we achieve 7.5 AP gains and set new state-of-the-art results among other monocular methods. Code will be available at: \url{https://github.com/WXinlong/ForeSeE}.
\end{abstract}

\section{Introduction}

Depth bridges the gap between 2D and 3D perception in computer vision. 
A precise depth map of an image provides rich 3D geometry information like locations and shapes for objects and stuff in a scene, thus attracting more and more attention in both 2D and 3D understanding fields.
Monocular depth estimation, which aims to predict the depth map from a single image, is an ill-posed problem, 
as infinite number of 3D scenes can be projected to the same 2D image.
With the development of deep convolutional neural networks~\cite{krizhevsky2012imagenet,simonyan2014very,he2016deep},
recent works have made great progress~\cite{XuW2018,FuGWBT18,li2018deep}.
They typically consist of an encoder for feature extraction and a decoder for generating the depth of the whole scene, either by regressing the depth values or predicting the depth range categories.
Plausible results have been shown.
%

\begin{figure}[!tb]
\centering
\includegraphics[width=0.45\textwidth]{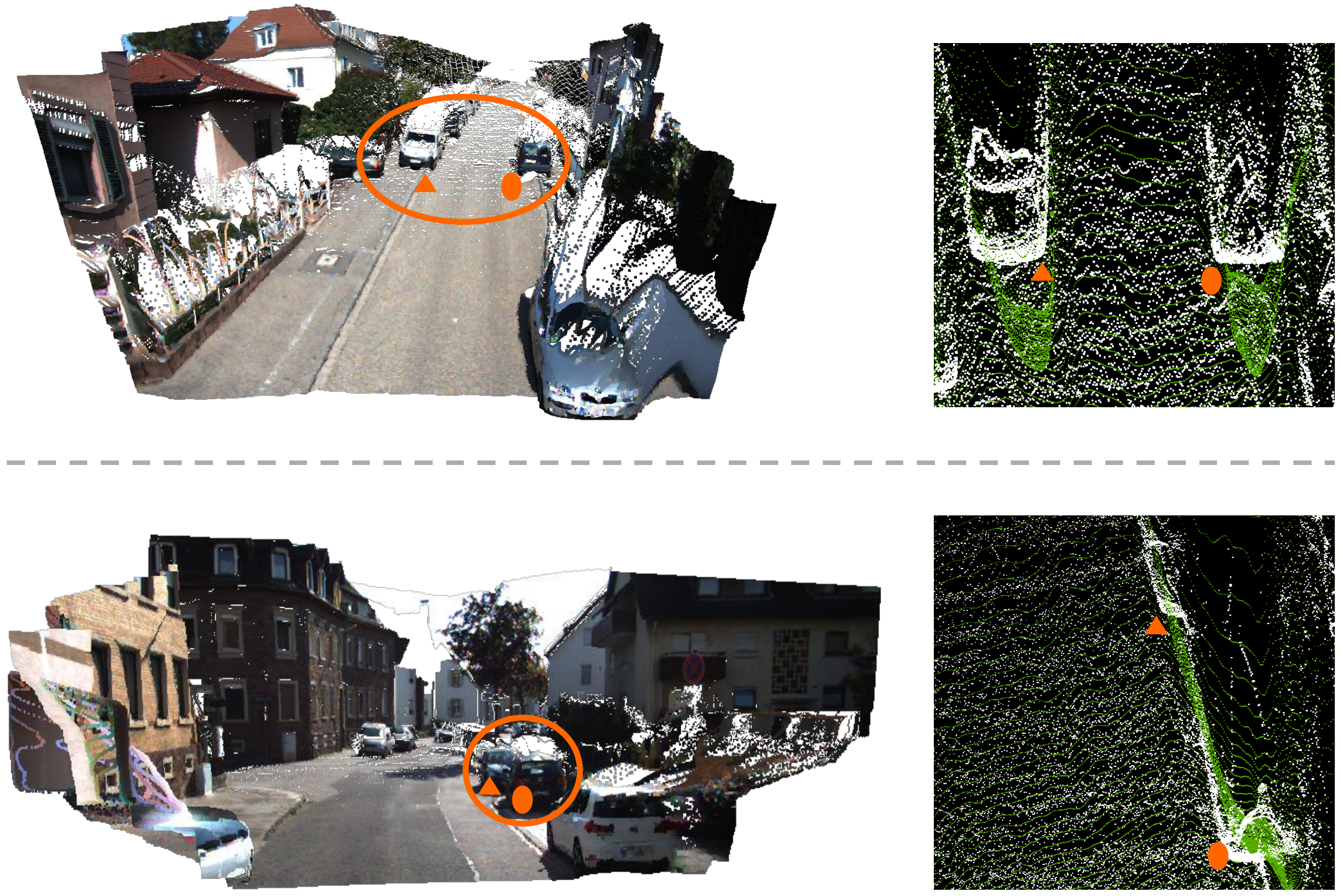}
\caption{ Examples of low precision prediction of foreground depth. For each row, the left picture is the projected point cloud transformed from ground truth depth map and RGB image; the right picture is the bird's-eye-view close-up to compare the depth (in green) predicted by the baseline depth estimation method with the ground truth (in white). 
The inaccurate object location and shape pose challenges for 3D recognition, localization and orientation estimation.
}
\label{fig:intro}
\end{figure}

When the monocular methods are applied 
to other tasks focusing on foreground object analysis, \eg, 3D object detection, there are two main obstacles from the low precision of foreground depth:
(1) Poor estimate of the object center location;
(2) Distorted or faint object shapes.
We show some examples in Figure~\ref{fig:intro}.
The inaccurate object location and shape make the downstream localization and recognition challenging.
The above issues could be handled by enhancing the depth estimation performance on foreground regions.
However, all these state-of-the-art methods treat foreground depth and background depth equally, which leads to sub-optimal performance on foreground objects.

\begin{figure}[!tb]
\centering
\includegraphics[width=0.41\textwidth]{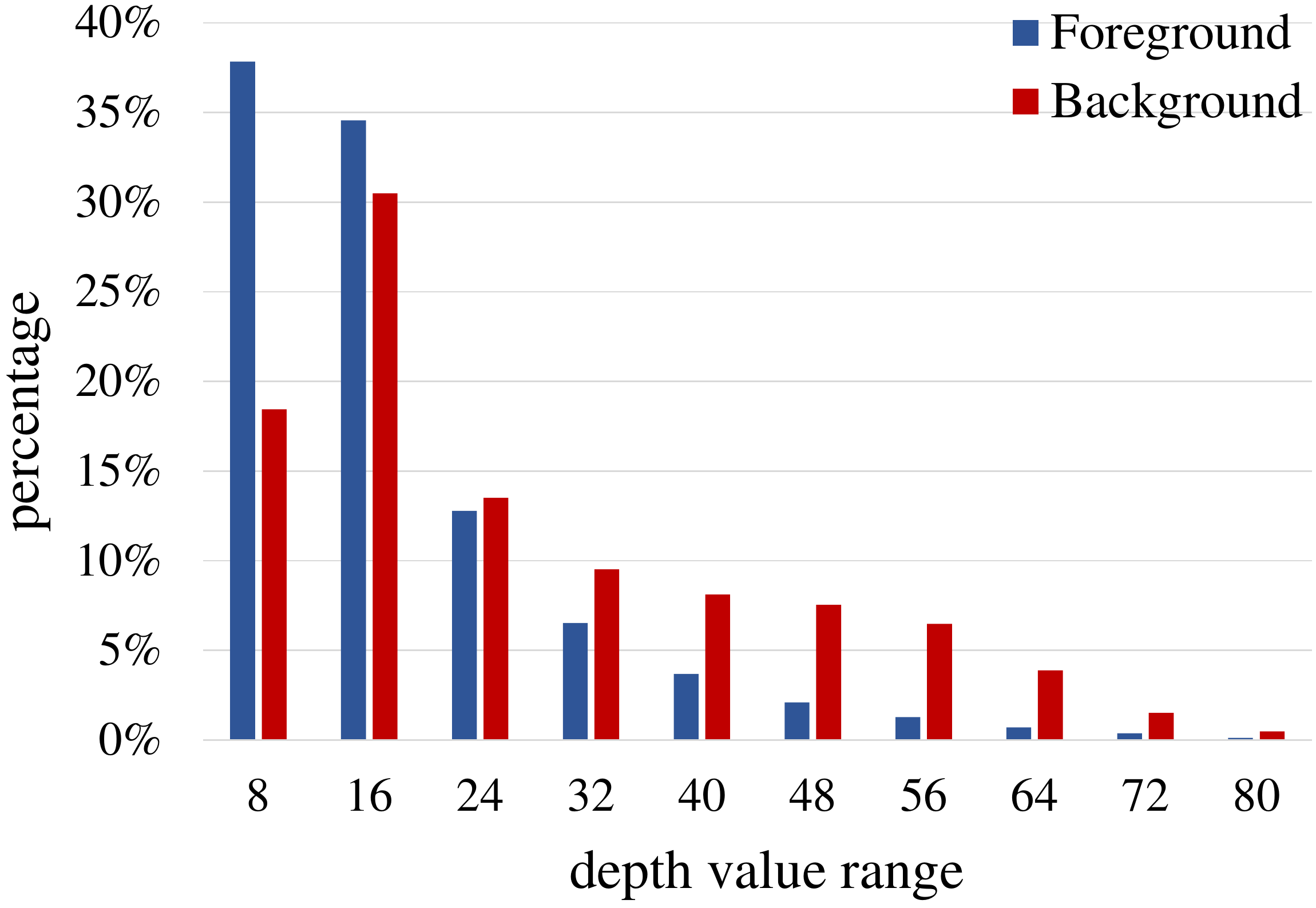}
\caption{Comparison of depth value distribution between foreground pixels and background pixels. Percentage of pixels with depth value within [$x$-8, $x$] meters is reported.}
\label{fig:data_dist}
\end{figure}

In fact, foreground depth and background depth show different data distributions.
We make qualitative and quantitative comparisons in Figure~\ref{fig:intro}, Figure~\ref{fig:data_dist} and Table~\ref{tab:depth_grad}. 
Foreground pixels tend to gather into clusters, bring more and bigger depth change and look like frustums in 3D space rather than flat surfaces like road and buildings.
Second, foreground pixels account for only a small part of the whole scene.
For instance, in the KITTI-Object dataset~\cite{GeigerLSU13}, $90.6\%$ pixels belong to background, while only $9.4\%$ pixels fall within foreground.
Furthermore, not all pixels are equal. As just described, foreground pixels play a more crucial role in downstream applications, \eg, autonomous driving and robotic grasping.
For example, an estimation error on a car is much different from the same error on a building. 
The inaccurate shape and location of the car could be catastrophic for 3D object detectors.

The observations make one wonder how to boost the estimation accuracy of foreground while do no harm to background.
First of all, it is neither a hard example mining problem, nor a self-learned local attention problem.
Different from the former one, here we want to further enhance the performance on specific regions in the scene, which does not have to be harder example.
As we can see in Figure~\ref{fig:fg-bg-weight}, foreground is indeed not harder than background.
Attention mechanism is widely used to focus on more discriminative local regions in semantic classification problems, \eg, semantic segmentation and fine-grained classification.
But this is not the case for depth estimation.
Given a close-up of a car in a scene, one could classify the semantic categories, but could not tell the depth.
Another choice is to separately train the foreground and background regions, since the data distributions are different. However, we show that the foreground and background are interdependent to each other for inferring the depth and boosting the performance. 

Instead, we formulate it as a multi-objective optimization problem. 
The objective functions of foreground and background depth are separated.
So do the depth decoders.
Thus, the foreground depth decoder could fit the foreground depth as well as possible while do no harm to background.

To summarize, our contributions are as follows:
\begin{itemize}
\setlength{\itemsep}{0pt}
\setlength{\parskip}{0pt}
\setlength{\parsep}{0pt}
  \item We conduct pioneering discussion about difference and interaction of foreground and background in monocular depth estimation.  We show that the different patterns of foreground and background depth lead to sub-optimal results on foreground pixels.
  \item We propose ForeSeE, to learn and predict foreground and background depth separately. Specifically, it contains separate depth decoders for foreground and background regions, 
  an objective sensitive loss funcion to optimize corresponding decoders,
  and a simple yet effective foreground-background merging strategy.
  \item With the proposed ForeSeE, we are able to predict much superior foreground depth, whereas background depth is not affected. 
  Furthermore, utilizing the predicted depth maps, our model achieves 7.5 AP gains on 3D object detection task, which effectively verifies our motivation.
\end{itemize}

\section{Related Work}
\myparagraph{Monocular Depth Estimation.}
Monocular depth estimation (MDE) is a long-lasting problem in computer vision and robotics. Early works~\cite{SaxenaSN09,SaxenaCN05} mainly leverage non-parametric optimization to predict the depth from handcrafted features~\cite{HoiemEH07,LadickySP14}. Recent powerful deep convolutional neural networks (DCNN) boost the performance of MDE significantly. Most methods formulate MDE as a pixel-wise supervised learning problem. Eigen~\etal ~\cite{EigenPF14} is the first to utilize the multi-scale DCNN to regress the depth map from a single image. Then, various innovative network architectures~\cite{LiuSL016,LiKY17,XuROWS19,FuGWBT18} are proposed to leverage strong high-level features.  Furthermore, several methods~\cite{zhaoFGT2019,QiLLUJ18,weiLSY2019} propose to explicitly enforce geometric constraints for the optimizing process. In this work, we focus on boosting the depth prediction accuracy of foreground objects with the proposed ForeSeE optimization strategy.

\myparagraph{Not All Pixels are Equal.}
Some prior works noticed that it is sub-optimal to treat all pixels equally in dense prediction tasks.
Sevilla~\etal~\cite{Sevilla-LaraSJB16} tackle optic flow estimation by defining different models of image motion for different regions.
Li~\etal~\cite{LiLLLT17} use deep layer cascade to first segment the easy pixels then the harder ones.
Sun~\etal~\cite{Sun_2019_CVPR} select and weight synthetic pixels which are similar with real ones for learning semantic segmentation.
Yuan~\etal~\cite{Yuan_2019_CVPR} introduce an instance-level adversarial loss for video frame interpolation problem.  
Shen~\etal~\cite{Shen_2019_CVPR} propose an instance-aware image-to-image translation framework.
However, different from the above works, we focus on depth estimation problem and aim at improving the accuracy of 3D object detection.

\myparagraph{Monocular 3D Object Detection.}
The lack of depth information poses a substantial challenge for estimating 3D bounding boxes from a single image.
Many works seek help from geometry priors and estimated depth information.
Deep3DBox~\cite{MousavianAFK17} proposes to generate 3D proposals based on 2D-3D bounding box consistency constraint.
ROI-10D~\cite{Manhardt_2019_CVPR} introduces RoI lifting to extract fused feature maps from input image and estimated depth map, before the 3D bounding box regression.
MonoGRNet~\cite{QinWL19} estimates the depth of the targeting 3D bounding box's center to aid the 3D localization.
Recently, some works~\cite{XuC18,Wang_2019_CVPR,wengK2019} propose to convert estimated depth map to lidar-like point cloud  to help localize 3D objects. 
Wang~\etal~\cite{Wang_2019_CVPR} directly applies 3D object detection methods on the generated pseudo-lidar, and claim 3D point cloud is a much superior representation than 2D depth map for better utilizing depth information. 
In these methods, a reliable depth map, especially the precise foreground depth, is the key to a successful 3D object detection framework.
We perform 3D object detection using the pseudo-LiDAR generated by our depth estimation model. 
The proposed method largely improves the performance and outperforms state-of-the-art methods.

\begin{table}[!thb]
\begin{center}
\small 
\setlength{\tabcolsep}{3.8pt}
\begin{tabular}{c|c|c|c}
\hline
 & I & II & III \\
 \hline
Foreground & 96.77 & 1.99 & 1.24\\
Background & 98.63 & 0.94 & 0.43  \\
\hline
\end{tabular}
\end{center}
\caption{Comparison of depth gradient distribution between foreground and background pixels. 
The gradients are uniformly discretized into three bins: I, II and III, from  small to large. Percentage of pixels at each level is reported.
}
\label{tab:depth_grad}
\end{table}

\section{Observation and Analysis}
\label{sec:analysis}

\subsection{Preliminaries}
\label{subsec:preliminaries}

\myparagraph{Dataset.}
KITTI dataset~\cite{GeigerLSU13} has witnessed inspiring progress in the field of depth estimation. 
As most of scenes in KITTI-Raw data have limited foreground objects, we construct a new benchmark which is based on KITTI-Object dataset.
We collect the corresponding ground-truth depth map for each image in KITTI-Object training set, and term it as KITTI-Object-Depth (KOD) dataset.
A total of $7,481$ image-depth pairs are divided into training and testing subsets with $3,712$ and $3,769$ samples respectively ~\cite{ChenKZBMFU15}, which makes sure that images in the two subsets belong to different video clips.
2D bounding boxes are used to distinguish the foreground and background pixels.
Pixels fall within 
the foreground bounding boxes are designated as foreground pixels, while the other pixels are assigned to be background.

\myparagraph{Baseline Method.} 
We adopt the same DCNN-based baseline method~\cite{weiLSY2019} which has already shown state-of-the-art performance on several benchmarks.
The main structure falls into the typical encoder-decoder style. 
Given an input image, the encoder extracts the dense features, then the decoder fetches the features and predicts the quantized depth range categories.
Specifically, the depth values are discretized into $100$ discrete bins in the log space. 
The quantized labels are assigned to each of the pixels as their classification labels.

\begin{figure}[!thb]
\centering
\includegraphics[width=0.41\textwidth]{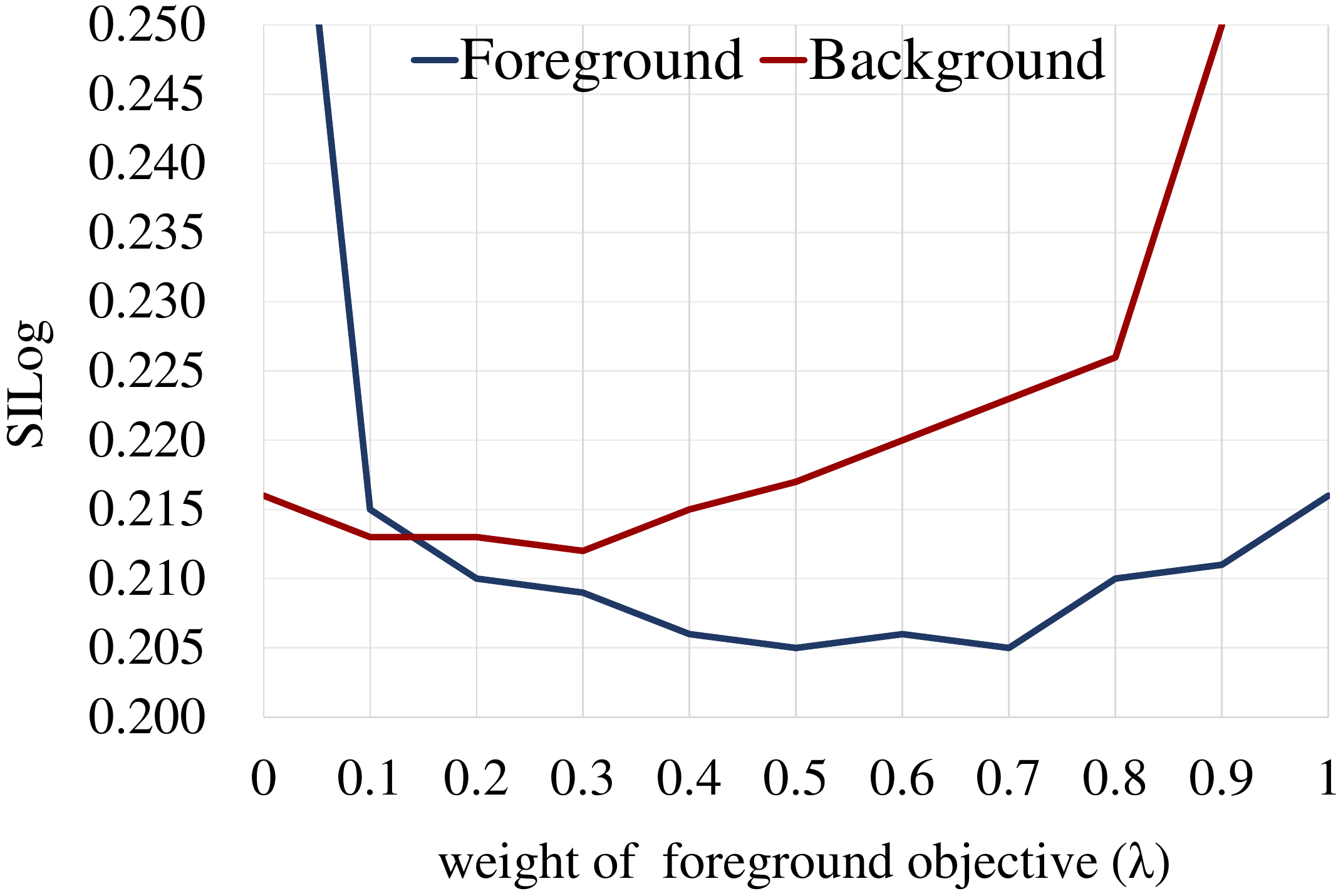}
\caption{Interaction of foreground and background samples. The depth estimation results (SILog) on foreground and background regions are reported (lower is better). The weight of foreground objective is on $x$-axis.}
\label{fig:fg-bg-weight}
\end{figure}

\begin{figure*}[htbp]
\centering
\includegraphics[width=0.95\textwidth]{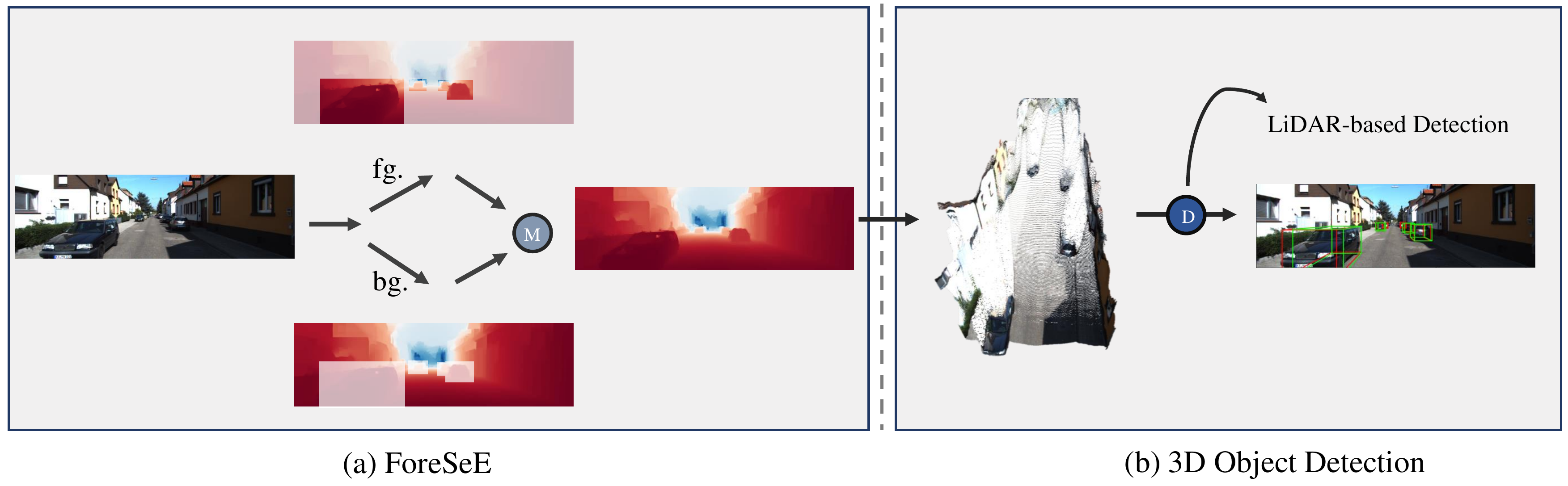}
\caption{Illustration of the overall pipeline.  (a) Foreground-background separated depth estimation. (b) 3D object detection.}
\label{fig:main}
\end{figure*}

\subsection{Analysis on Data Distribution}
\label{subsec:data_distribution}

Few works~\cite{Jiao_2018_ECCV} have analysed the depth distribution, not to mention the foreground and background depth distributions. 
Here we investigate two kinds of data distribution of foreground and background pixels in training subset. 
Figure~\ref{fig:data_dist} shows the depth value distributions.
As shown, more than $75\%$ foreground pixels have depth less than $16m$, while it is about $50\%$ for background.
The foreground depth also shows a heavier long-tail distribution.
Depth gradient distributions are shown in Table~\ref{tab:depth_grad}.
We use the Laplacian of the depth images as the depth gradient, which calculates the second order spacial derivatives.
The Laplacian image highlights areas of rapid depth change.
The outputs are scaled to [0, 255] and uniformly discretized into three bins: $\rm\uppercase\expandafter{\romannumeral1}$, $\rm\uppercase\expandafter{\romannumeral2}$ and $\rm\uppercase\expandafter{\romannumeral3}$, from small to large.
In this way, all pixels are divided into three levels according to their gradient values.
The foreground has much higher proportion than background at level $\rm\uppercase\expandafter{\romannumeral2}$ and $\rm\uppercase\expandafter{\romannumeral3}$.
Besides the depth range and depth gradient, the difference of shapes should also be noted.
Generally, depth provides two kinds of information: location and shape. 
The foreground objects share similar shapes and look like frustums in 3D space, as shown in Figure~\ref{fig:intro}. 
Based on the above analysis, we propose to consider the foreground and background separately when estimate their depth.

\subsection{Separate Objectives}
\label{subsec:separate_objectives}

In dense prediction tasks, generally the loss function can be formulated as:
\begin{equation}
L = \frac{1}{N} \sum_{i}^{N}E(y_{i}, \hat{y_{i}}),
\end{equation}
where $N$ is the number of pixels, $y_i$ and $\hat{y_{i}}$ are the prediction and ground truth of $i^{th}$ pixel.
$E$ is  the error function, \eg, the widely used cross-entropy error function.

After the analysis in Section~3.2, we further investigate the interaction of foreground and background by splitting the optimization objective.
The modified loss function is defined as:
\begin{equation}
L = \lambda \times \frac{1}{N_f} \sum_{i}^{N_f}E(y_{i}, \hat{y_{i}}) + (1 - \lambda) \times \frac{1}{N_b} \sum_{i}^{N_b}E(y_{i}, \hat{y_{i}}),
\end{equation}
where $N_f$ is the number of foreground pixels, $N_b$ is the number of background pixels and $\lambda$ acts as the weight to balance the two loss terms.
Figure~\ref{fig:fg-bg-weight} shows our results with different settings of $\lambda$. The results are generated by a CNN that has a single depth prediction decoder but the separated objective function.
When $\lambda$ is set to $0$, which means only the background samples are used to supervise the training, the result on foreground becomes very poor.
Similarly, the performance on background drops sharply when $\lambda$ is set to $1.0$.
It verifies that the foreground depth and background depth are distributed differently.
When we increase the foreground weight $\lambda$ from $0$ to $0.1$, the result on background improves, which indicates that the foreground and background to some extend could help each other. 
Further, it should be noted that the optimal $\lambda$ values for foreground and background are different.
For instance, the model shows its best performance on foreground when $\lambda=0.7$, but meanwhile the result on background is much poorer.
It indicates that the optimization objectives for foreground and background are not consistent.
To address these issues, in Section~4, the foreground-background separated depth estimation method is proposed to
achieve the optimum points at the same time.

\subsection{Analysis Summary}
\label{subsec:conclusion}

We highlight three observations:
\begin{itemize}
    \item The foreground and background depth have different depth value distributions, depth gradient distributions and shape patterns;
    \item The foreground and background depth reinforce each other due to their shared similarities;
    \item The optimization objectives of foreground and background depth estimation are mismatched.
\end{itemize}

\begin{table*}[!thb]
\small 
\begin{center}
\begin{tabular}{c|ccc|cc|cc|cc}
\hline
\hline
    \multirow{2}{*}{Method} & \multirow{2}{*}{FSL}  & \multirow{2}{*}{SD} &  \multirow{2}{*}{SO} &  \multicolumn{2}{c|}{Foreground}  & \multicolumn{2}{c|}{Background}  & \multicolumn{2}{c}{Global} \\
\cline{5-10}
& & & & absRel & SILog & absRel & SILog & absRel & SILog \\
\hline
\hline
ForeSeE   & \checkmark & \checkmark & \checkmark & \bf0.118 & \bf0.205  & \bf0.141 & 0.210 & \bf0.138  & 0.210 \\
\hline
 & & \checkmark & \checkmark & 0.120 & 0.208 & \bf0.141 & \bf0.209 & 0.139 & \bf0.209 \\
 \hline 
 & & & \checkmark & 0.120 & \bf0.205 & 0.147 & 0.217 & 0.144 & 0.216 \\
\hline
Baseline & & & & 0.129 & 0.216  & 0.143 & 0.210 & 0.141 & 0.211 \\
\hline 
\end{tabular}
\end{center}
\caption{Ablation study of depth estimation on the KOD dataset. FSL refers to foreground-background sensitive loss; SD refers to separate decoders; SO means separate objectives. }
\label{tab:depth_ablation}
\end{table*}

\section{ForeSeE}
\label{sec:method}

In this section we first introduce the network architecture of our method, then present the proposed loss function, 
and finally show how the mask used to distinguish foreground and background could be dropped during the inference.
The whole pipeline is illustrated in Figure~\ref{fig:main}.

\subsection{Separate Depth Decoders}
\label{subsec:sdd}
We construct an additional decoder based on the baseline method~\cite{weiLSY2019}, thus there are two parallel decoders which have the same structure. 
One of the decoders is for foreground depth prediction, while the other one aims to estimate the background depth.
Specifically, for an image of size $H \times W \times 3$, each decoder outputs a tensor of size $H \times W \times C$, where $C$ is the number of depth range categories.

Foreground regions are cropped from the output of foreground depth decoder.
The background depth range predictions are obtained in the same way.
The global depth range predictions are generated by a seamless merge of foreground and background regions.
Then the depth range predictions are converted to the final depth map using the soft-weighted-sum strategy~\cite{LiDH18}. 

\subsection{Foreground-background Sensitive Loss Function}
As observed in Section~3, although the foreground depth and background depth show different patterns, they do share some similarities and could reinforce each other under an appropriate ratio.
Thus, we further weight the foreground and background samples.
For either foreground or background branch, the loss function is a weighted average of foreground samples and background samples, but with different bias.
Here we define the loss function which supervises the foreground branch as:
\begin{equation}
L_{fg} = \lambda_f  \times {E}_{fg} + (1 - \lambda_f) \times {E}_{bg},
\end{equation}
where ${E}_{fg}$ represents mean errors calculated on foreground predictions; ${E}_{bg}$ is the mean error of foreground predictions; $\lambda_f$ is the weight to balance the foreground and background samples during the training of foreground branch.
Larger $\lambda_f$ leads to more preference for foreground samples.
Similarly, the loss function of background branch is formulated as:
\begin{equation}
L_{bg} = \lambda_b \times {{E}{^ \prime}_{bg}} + (1 - \lambda_b) \times {{E}{^ \prime}_{fg}},
\end{equation}
where $\lambda_b$ is the weight; ${{E}{^ \prime}_{bg}}$ and ${{E}{^ \prime}_{fg}}$ are the mean errors of background predictions and foreground predictions on this background branch.

\subsection{Inference without Mask}

Here we propose a mask-free merge method such that the binary mask is no longer needed once the training is finished.
A max-pooling operation is applied on the foreground and background outputs before the softmax operation, which represent the confidence scores of being each range category. 
For each range category of each pixel, the highest confidence score between foreground and background output is retained, to serve as the final prediction. Formally, for the $H$$\times$$W$$\times$$C$ shaped outputs $P$ $=$ $p_1$,$ p_2$, $...$, $p_{H \times W}$ ($p_i \subseteq \mathbb{R}^{C}$),  
The final predictions are calculated as:
\begin{equation}
p_{i}^{\prime} = {\rm Max}(p_{i}^{f}, p_{i}^{b}),
\end{equation}
where $p_{i}^{f}, p_{i}^{b}$ represent the $i^{th}$ output of foreground and background branch, and ${\rm Max(\cdot)}$ is an element-wise maximum operator which takes two vectors as input and outputs a new vector. 
Then the $H$$\times$$W$$\times$$C$ shaped output is fed to $softmax$ and soft-weighted-sum operations to produce the final depth map.
The results only drop slightly compared with the mask-based merge method (from $0.117$ to $0.118$ absRel).

\section{Experiments}

\subsection{Experiment Settings}
\label{subsec:exp_settings}

\myparagraph{Datasets.}
We carry out experiments on KITTI dataset, which contains large-scale road scenes captured on driving cars, and serves as a popular benchmark for many computer vision problems related to self-driving cars. 
Specifically, we construct the KITTI-Object-Depth (KOD) dataset for evaluating the foreground depth estimation, as described in Section~3.1.
The KOD dataset will be public available for convenience of future researches. 
Besides, we also apply our method on KITTI-Object dataset to perform monocular 3D object detection.

\myparagraph{Evaluation Metrics.}
For evaluation of depth estimation, we follow common practice~\cite{li2018deep,weiLSY2019} and use the mean absolute relative error (absRel) and  scale invariant logarithmic error (SILog) as the main metrics. We also report mean relative squared error (sqRel), mean $\log_{10}$ error ($\log_{10}$) and accuracy under threshold ($\delta_i$). 
As for 3D object detection, we follow the prior works~\cite{Liu_2019_CVPR,Qin_2019_CVPR} and focus on the ``car'' class. We report the results of 3D and bird's-eye-view (BEV) object detection on the validation set. The commonly used average precision (AP) with the IoU thresholds at 0.7 is calculated.
The results on KITTI easy, moderate and hard difficulty levels are reported.

\myparagraph{Implementation Details.}
For depth estimation, we follow the most of settings in baseline method~\cite{weiLSY2019}. 
The ImageNet pretrained ResNeXt-101~\cite{XieGDTH17} is used as the backbone model.
We train the network for $20$ epochs, with batch size $4$ and base learning rate set to $0.001$.
The  Stochastic Gradient Descent (SGD) solver is adopted to optimize the network on a single GPU. 
$\lambda_f$ and $\lambda_b$ in foreground-background sensitive loss function are set to $0.2$.
Given a predicted depth map, the point cloud can be reconstructed based on the pinhole camera model. 
We transform each pixel $(u_i, v_i)$ with depth value $ d_i $ to a 3D point $(x_i, y_i, z_i)$ in left camera coordinate as follows:
\begin{equation}
z_i = d_i,
\end{equation}
\begin{equation}
x_i = \frac{d_i \times (u_i - c_U)}{f_U},
\end{equation}
\begin{equation}
y_i = \frac{d_i \times (v_i - c_V)}{f_V},
\end{equation}
where $f_U$ and $f_V$ are the focal length along the $x$ and $y$ coordinate axis; $c_U$ and $c_V$ are the 2D coordinate of the optical center.
Following~\cite{Wang_2019_CVPR}, we set the reflectance to $1$ for each point and remove the points higher than 1 $m$ above the LiDAR source.
The resulting point cloud is termed as pseudo-LiDAR.
Afterwards, any existing LiDAR-based 3D object detection methods can be applied.

\begin{table*}[!thb]
\begin{center}
\small 
\setlength{\tabcolsep}{3.8pt}
\begin{tabular}{c|c|c|c|c|c|c|c|c}
\hline 
\hline
  & Method    	& absRel $\downarrow$   & sqRel $\downarrow$	& SILog $\downarrow$  &	log10 $\downarrow$	& $\delta_1$ $\uparrow$ & $\delta_2$ $\uparrow$ & $\delta_3$ $\uparrow$     \\
\hline
\hline
\multirow{2}{*}{Foreground} 
  & DenseDepth & 0.135 & 0.214 & {\bf 0.204} & 0.057 & 0.830 & 0.951 & {\bf 0.984}  \\ 
  & ForeSeE 		& 	{\bf 0.118}	&  {\bf 0.193}	& 0.205 &  {\bf 0.053}	& {\bf 0.851} 	& {\bf 0.952} & 0.982   	\\
\hline
  \multirow{2}{*}{Global}  
  & DenseDepth & \bf0.138 & \bf0.208  & \bf0.209 &  0.062 & 0.782 & 0.946 & \bf0.987  \\  
  & ForeSeE &  \bf0.138  &  0.213  & 0.210 &  \bf0.061  & \bf0.793 & \bf0.949 & \bf0.987  \\

\hline
\end{tabular}
\end{center}
\caption{Depth estimation performance comapred with  DenseDepth~\cite{AlhashimW2018}.}
\label{tab:depth_results}
\end{table*}

\subsection{Depth Estimation}

\myparagraph{Quantitative Results.}
We show the ablative results in Table~\ref{tab:depth_ablation}.
Our ForeSeE outperforms the baseline over all metrics evaluated on foreground, background and global levels. 
Specifically, when evaluate on foreground level, our method improves the baseline performance by up to $ 8.5\% $ (from $0.129$ to $0.118$ absRel).
It is in accordance with our intention that ForeSeE is specifically designed to enhance the ability of estimating the foreground depth. 
We further analyse the effect of each component.
When equipped with the separate objectives (SO) described in Section~3.3, the baseline achieves better results on foreground while performs worse when evaluate on background pixels.
Directly using the separate decoders (SD) could avoid the harm on background.
Finally, the performance on foreground is further improved by applying the foreground-background sensitive loss (FSL).

To compare with other state-of-the-art methods, we apply DenseDepth~\cite{AlhashimW2018} to the KOD benchmark, which reports the best performance on KITTI~\cite{GeigerLSU13} and NYUv2~\cite{SilbermanHKF12} datasets among the methods with public available training code.
We obtain results of DenseDepth using the code at github\footnote{https://github.com/ialhashim/DenseDepth}, published by the authors.
Except the dataset used, all the settings and hyper-parameters are not modified. 
The results are shown in Table~\ref{tab:depth_results}.
Compared with DenseDepth, our ForeSeE shows significant advantage on foreground level.
For instance, ForeSeE outperforms DenseDepth by absolute $ 2.7\% $ absRel (from $0.135$ to $0.118$ absRel), which is a relative improvement of $12.6\%$.

\begin{figure}[!thb]
\centering
\includegraphics[width=0.45\textwidth]{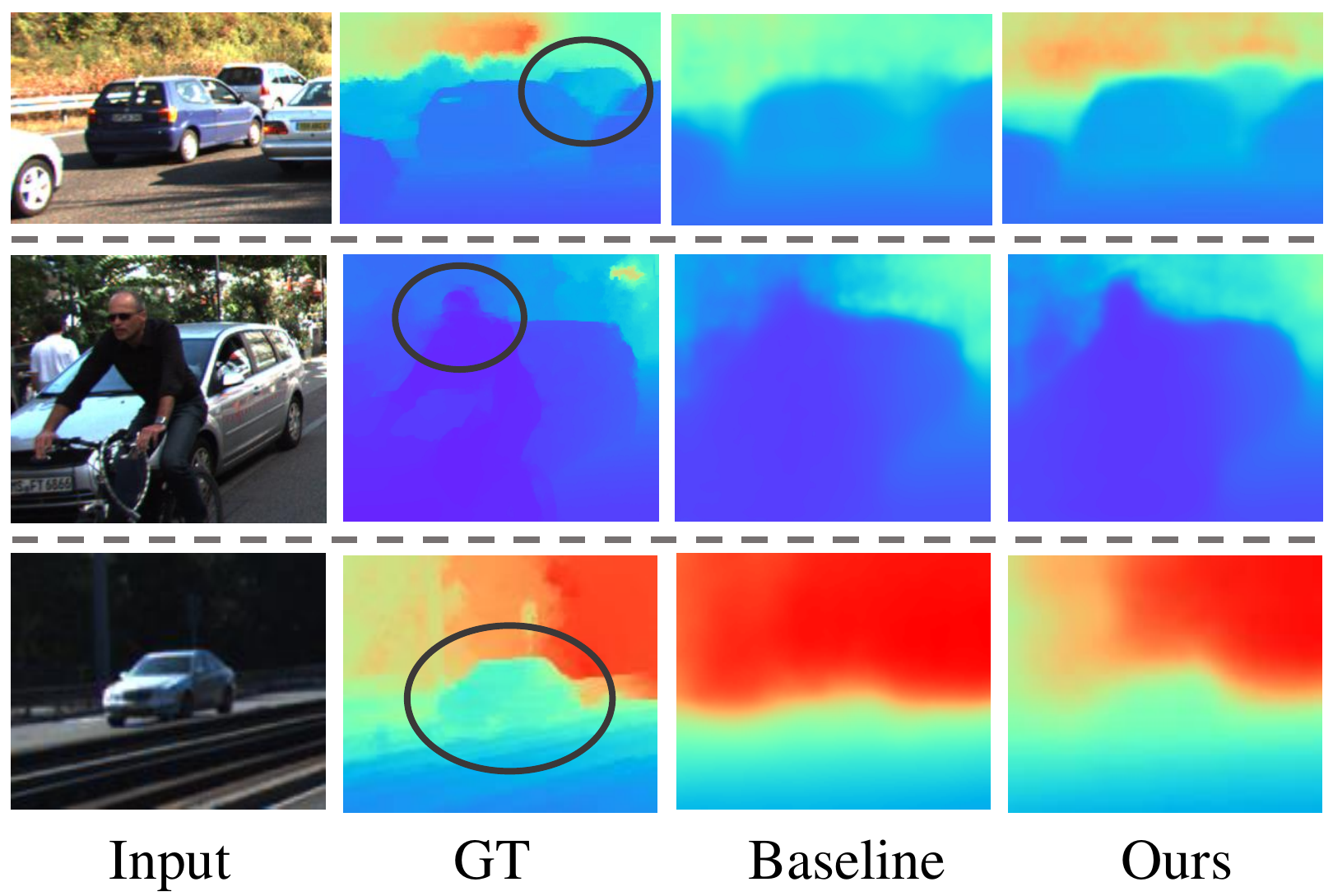}
\caption{Quantitative comparison of the baseline and our ForeSeE on estimated depth maps.}
\label{fig:exp_vis_depth}
\end{figure}

\begin{figure}[!thb]
\centering
\includegraphics[width=0.45\textwidth]{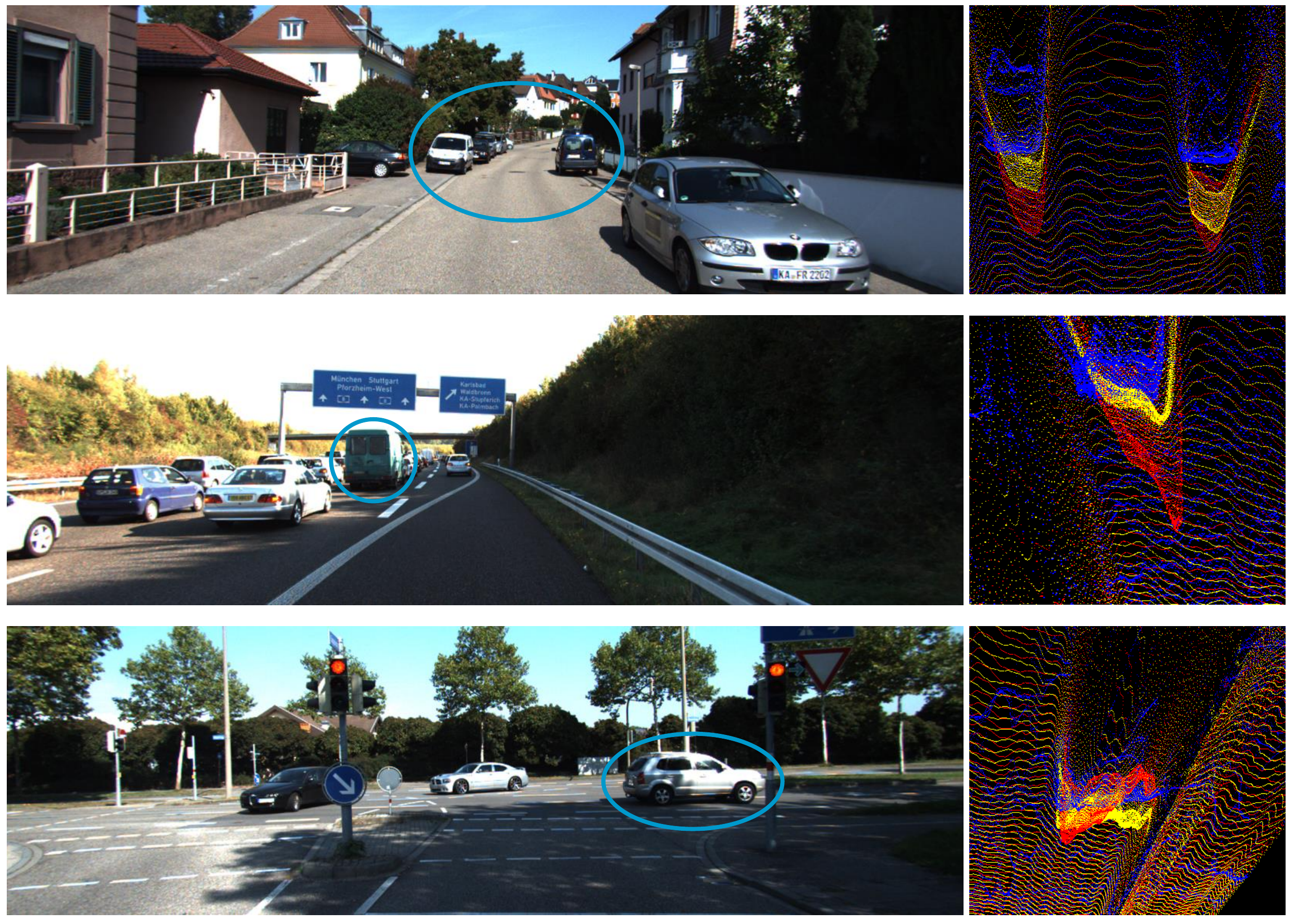}
\caption{Quantitative comparison of the baseline and our ForeSeE on converted pseudo-LiDAR signals. Signals in blue are converted from ground truth depth; Baseline pseudo-LiDAR are in red; Our ForeSeE pseudo-LiDAR are in yellow.}
\label{fig:exp_vis_pc}
\end{figure}

\myparagraph{Qualitative Results.}
Besides the quantitative comparison, we show some visualization results.
The predicted depth maps are visualized in Figure~\ref{fig:exp_vis_depth}.
As shown, our ForeSeE estimates more precise depth on foreground regions.
The contour of foreground objects is more clear and accurate.
Further, in Figure~\ref{fig:exp_vis_pc} we compare the estimated depth in the format of 3D point cloud.
3D point cloud is a more intuitive and reasonable representation for visually comparing and debugging depth maps.
As shown in Figure~\ref{fig:exp_vis_pc}, our method shows less estimation errors and more accurate bird's-eye-view (BEV) shapes.

\begin{figure*}[!tb]
\centering
\includegraphics[width=0.97\textwidth]{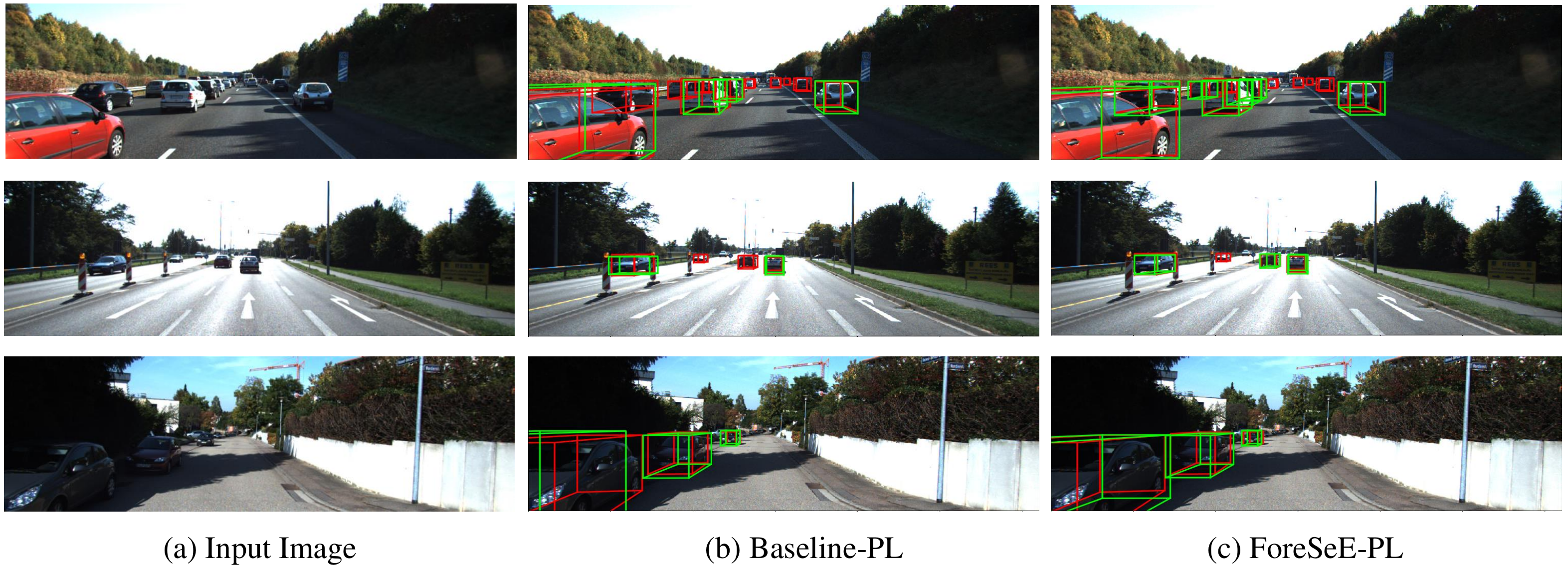}
\caption{Qualitative results of 3D object detection. The ground truth 3D bounding boxes are in red; the predictions are in green.}
\label{fig:3ddet_vis}
\end{figure*}

\subsection{3D Object Detection}

To further validate the effectiveness, we conduct experiments on 3D object detection problem.
We convert the estimated image-based depth map to LiDAR-like point cloud (pseudo-LiDAR).
Then the LiDAR-based algorithms can be applied to recognizing and localizing 3D objects.
Here we adopt Frustum-PointNet (F-PointNet)~\cite{QiLWSG18} and AVOD~\cite{KuMLHW18}, specifically the F-PointNet-v1 and AVOD-FPN, which are top-performing 3D object detection methods and both utilize the information from LiDAR and RGB images.

\myparagraph{Brief Introduction of Detection Methods.}
Frustum-PointNet leverages 2D detector to generate 2D object region proposals in a RGB image.
Each 2D region corresponds to a 3D frustum in 3D space.
PointNet-based networks are used to estimate a 3D bounding box from the points within the frustum.
AVOD uses multimodal feature fusion RPN which aggregates the front-view image features and BEV LiDAR features to generate 3D object proposals.
Based on the proposals, the bounding box regression and category classification are performed in the second subnetwork. 
We apply the F-PointNet and AVOD on the pseudo-LiDAR generated by our depth estimation model during the training and inference. 
Hyper-parameters are not modified.
More details about the 3D object detector can be referred to the original papers.

\myparagraph{Comparisons with State-of-the-art Methods.}
It should be noted that some works~\cite{Wang_2019_CVPR,wengK2019} use the DORN~\cite{FuGWBT18} pre-trained on KITTI-raw dataset for depth estimation, which includes the images in training and validation subsets of KITTI-Object.  
Wang~\etal~\cite{Wang_2019_CVPR} claim that their results serve as the upper bound.
If we also pre-train our baseline depth estimation model on KITTI-raw and use it to generate pseudo-LiDAR, we achieve $20.1$ AP which outperforms their reported $18.5$ AP when both use the F-PointNet as detector and evaluate on the moderate level of car class.
But, we want to clearly set a baseline and fairly compare to other state-of-the-art monocular 3D detection methods.

We compare our method with other methods in Table~\ref{tab:3ddet}. 
The methods are evaluated by the average precision (AP) with IoU threshold at 0.7. All the methods are tested on ground-truth 3D bounding boxes. 
The compared 3D object detection methods include Mono3D~\cite{ChenKZMFU16}, MLF-MONO~\cite{XuC18}, ROI-10D~\cite{Manhardt_2019_CVPR}, MonoGRNet~\cite{QinWL19}, MonoPSR~\cite{Ku_2019_CVPR}, TLNet-Mono~\cite{Qin_2019_CVPR} and DFDSNet~\cite{Liu_2019_CVPR}.
Our depth estimation models are trained on KOD training subset which does not contain validation subset of KITTI-Object.
Either with F-PointNet or AVOD as the detection method, our ForeSeE-PL brings remarkable improvements on the basis of baseline-PL over all the metrics, \eg, from $19.0$ to $23.4$ $\rm{AP}_{BEV}$  with AVOD detector.
Note that the 3D detection average precision ($\rm{AP}_{3D}$) is the most widely used metric, on which our method achieves $7.5$ AP gains (from $7.5$ to $15.0$ AP) and outperforms all the state-of-the-art methods.
Another advantage of our method is that it is not limited to specific 3D object detection methods. 
With stronger 3D object detector, we achieve larger improvements, \eg, $3.6$ AP gains on F-PointNet and $7.5$ AP gains on AVOD when evaluate on easy level of $\rm{AP}_{3D}$.

\begin{table}[!ht]
\begin{center}
\footnotesize
\setlength{\tabcolsep}{3.6pt}
\begin{tabular}{c|c|c|c}
\hline
\hline
Method &   Easy & Moderate & Hard \\
\hline
\hline
Mono3D &    5.2 / 2.5  & 5.2 / 2.3  & 4.1 / 2.3   \\
MLF-MONO &   22.0 / 10.5  &  13.6 / 5.7   &  11.6 / 5.4  \\
ROI-10D &    14.5 / 9.6 & 9.9 / 6.6 & 8.7 / 6.3  \\
MonoGRNet &   ~~~-~~~  /  13.9 & ~~~-~~~ / 10.2 & ~~~-~~~ / 7.6  \\
MonoPSR~ &  20.6 / 12.8 & {\bf 18.7} / 11.5 & 14.5 / 8.6 \\
TLNet-Mono &  21.9 / 13.8  & 15.7 / 9.7 & 14.3 / 9.3 \\
DFDSNet &  9.5 / 6.0  & 8.0 / 5.5 & 7.7 / 4.8 \\
\hline	
\hline 
F-PN (baseline-PL)   &   17.3 / 9.6 & 11.8 / 5.4 & 10.4 / 5.0 \\
F-PN (Our ForeSeE-PL)   &  20.2 / 13.2 & 12.6 / 9.4 & 12.0 / 8.2 \\
\hline
AVOD (baseline-PL) &  19.0 / 7.5 & 15.3 / 6.1 & 13.0 / 5.4  \\
AVOD (Our ForeSeE-PL) &  {\bf 23.4} / \bf15.0 & 17.4 / \bf12.5 & {\bf 15.9} / \bf12.0 \\
\hline 
\end{tabular}
\end{center}
\caption{Monocular 3D object detection results on KITTI benchmark. We report $\rm{AP_{BEV} / AP_{3D}}$ (in \%)  of the car category. F-PN refers to Frustum-PointNet. PL refers to pseudo-LiDAR. Here ForeSeE-PL stands for using pseudo LiDAR from ForeSeE.}
\label{tab:3ddet}
\end{table}

\myparagraph{Qualitative Results.}
The visualization of detection results are shown in Figure~\ref{fig:3ddet_vis}.
The 3D bounding boxes are projected into image space for better visualization.
There are two obvious advantages of using ForeSeE-PL: less missed detection and more accurate localization.
Inaccurate depth predictions will result in shifted localization or rotated orientation, as in Figure~\ref{fig:3ddet_vis}(b).
Even worse, the objects can not be detected if the depth estimation model treats foreground objects as background region, thus causing more missed detections.
Our ForeSeE method largely alleviates the problems through predicting more accurate depth on foreground regions.

\section{Conclusion}
In this paper, we first analyse the data distribution of foreground and background depth and explicitly explore the interactions.
Based on the observations, a simple and effective depth estimation pipeline, namely ForeSeE, is proposed to estimate foreground depth and background depth separately.
We introduce a foreground depth estimation benchmark and set fair baselines to encourage the future studies.
The experiments on monocular depth estimation and 3D object detection problems demonstrate the effectiveness of ForeSeE.
We expect wide application of the proposed method in depth estimation and related downstream problems, \eg, 3D object recognition and localization.

{\small
\bibliographystyle{aaai}
\bibliography{main}
}

\end{document}